\documentclass[11pt]{article}

\usepackage[margin=1in]{geometry}
\usepackage{amsmath,amssymb}
\usepackage{array}
\usepackage{booktabs}
\usepackage{nicematrix}
\usepackage{graphicx}
\usepackage{subcaption}
\usepackage{algorithm}
\usepackage{algpseudocode}
\algrenewcommand\algorithmicrequire{\textbf{Input:}}
\algrenewcommand\algorithmicensure{\textbf{Output:}}
\usepackage{natbib}
\usepackage[hidelinks]{hyperref}

\title{CellRFT: Reinforcement Fine-Tuning for Single-Cell Perturbation Modeling}
\newcommand{\paperauthors}{%
  \begin{minipage}{\dimexpr\textwidth-2\tabcolsep\relax}
    \centering
    \normalsize
    Jie Yan\textsuperscript{1,\textdagger},
    Li Liu\textsuperscript{2,\textdagger},
    Hanze Guo\textsuperscript{3},
    Jiaxin Hu\textsuperscript{1},\\[3pt]
    Houxin He\textsuperscript{1},
    \href{https://orcid.org/0000-0003-3447-2842}{Xiaoning Qi}\textsuperscript{1},
    Haoran Wang\textsuperscript{1,4},\\[3pt]
    Cong Li\textsuperscript{1},
    Zhong-Yuan Zhang\textsuperscript{5},
    Yong Wang\textsuperscript{1,4,*}
    \par\vspace{9pt}
    \small
    \textsuperscript{1}Chinese Academy of Sciences\par
    \textsuperscript{2}Peking University\par
    \textsuperscript{3}Renmin University of China\par
    \textsuperscript{4}University of Chinese Academy of Sciences\par
    \textsuperscript{5}Central University of Finance and Economics
  \end{minipage}%
}
\author{\paperauthors}
\date{}

\begin{document}

\maketitle
\begingroup
\renewcommand{\thefootnote}{\fnsymbol{footnote}}
\footnotetext[1]{Corresponding author:
  \href{mailto:ywang@amss.ac.cn}{\texttt{ywang@amss.ac.cn}}}
\footnotetext[2]{These authors contributed equally to this work.}
\endgroup

\begin{abstract}
Predicting cellular responses to perturbations supports the study of gene
function, disease mechanisms, and therapeutic strategies.
Despite advances in single-cell perturbation modeling, existing models
typically optimize surrogate losses that do not directly reflect the
biological criteria used for evaluation, so better data fitting need not
yield better biological predictions.
To address this mismatch, we introduce \textbf{CellRFT}, a reinforcement
fine-tuning framework that uses biological evaluation as direct training
feedback.
CellRFT uses policy-gradient optimization to learn from non-differentiable
evaluations of generated cell populations and integrates multiple
biological rewards through hierarchical reward aggregation.
Comprehensive experiments demonstrate CellRFT's applicability across
different pretrained models and effectiveness in improving perturbation
prediction, reveal that optimizing one biological criterion can help or
hinder others, and show that complementary rewards can improve criteria
beyond those directly optimized, offering a way to probe how biological
metrics shape model behavior, with the potential to inform evaluation design.
Code will be made available.

\end{abstract}

\section{Introduction}
\label{sec:introduction}


Predicting cellular responses to genetic and chemical perturbations extends
experimental coverage beyond feasible
screens~\citep{bunne2024build,adduri2025predicting}, supporting the study
of gene function and disease mechanisms and the discovery of therapeutic
strategies~\citep{dixit2016perturb,bunne2024build}.
Perturbation models are typically trained with
reconstruction~\citep{lotfollahi2019scgen},
denoising~\citep{yuan2026perturbdiff}, or
distribution-matching~\citep{adduri2025predicting} losses, then evaluated on expression fidelity,
affected-gene recovery, and perturbation discrimination using suites such
as Cell-Eval~\citep{adduri2025predicting}.
These surrogate losses do not directly optimize the biological criteria,
creating a \textit{training--evaluation objective mismatch} in which better
data fitting need not improve biological prediction ~\citep{zhu2025auprc, ahlmann2025deep}. 

\begin{figure}[!htb]
  \centering
  \captionsetup[subfigure]{font=footnotesize,labelfont=bf,
    justification=raggedright,singlelinecheck=false,position=top,skip=2pt}
  \begin{subfigure}[t]{0.56\linewidth}
    \centering
    \caption{}
    \label{fig:intro-overview-pipeline}
    \includegraphics[width=\linewidth]{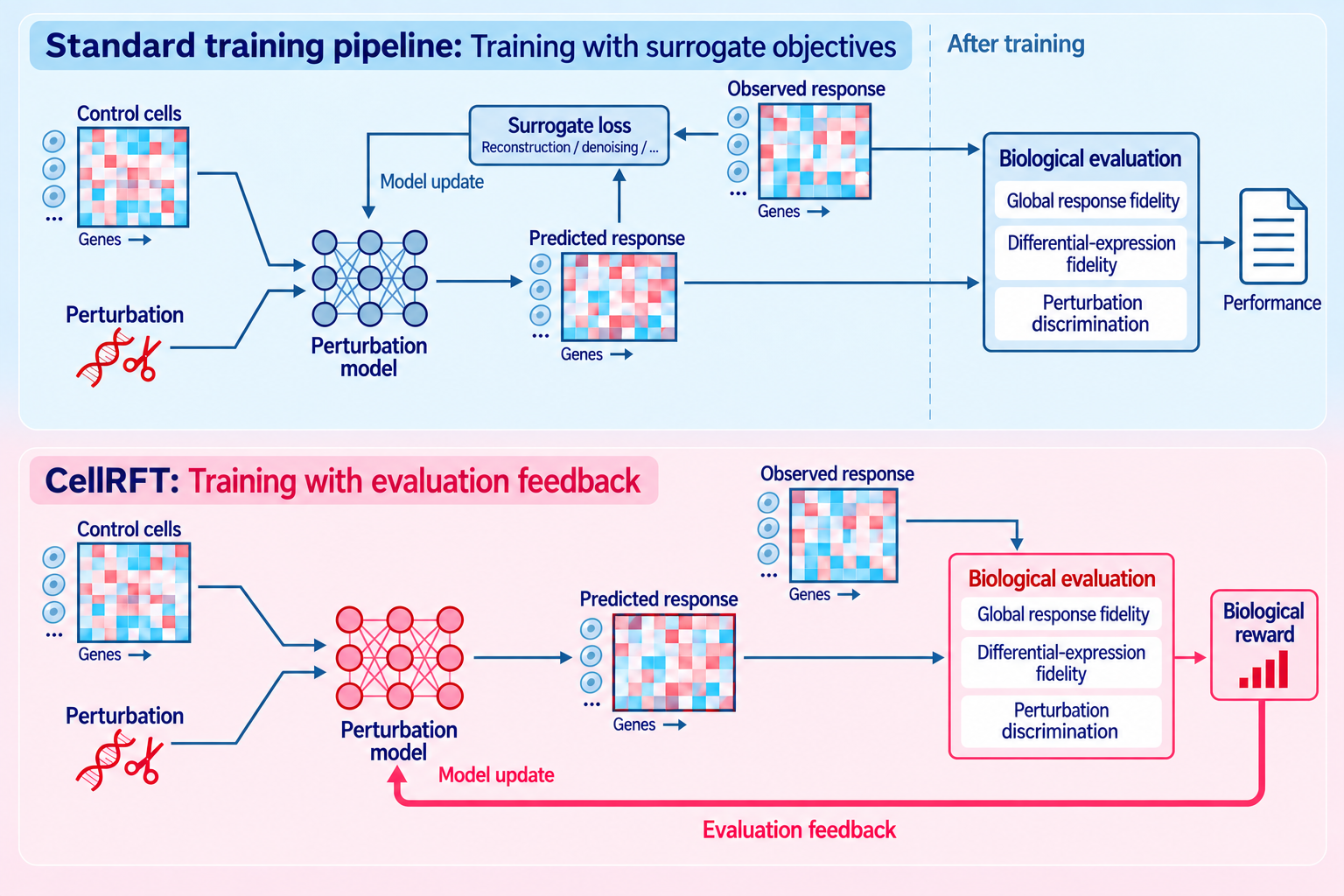}
  \end{subfigure}\hfill
  \begin{subfigure}[t]{0.42\linewidth}
    \centering
    \caption{}
    \label{fig:intro-overview-radar}
    \includegraphics[width=\linewidth]{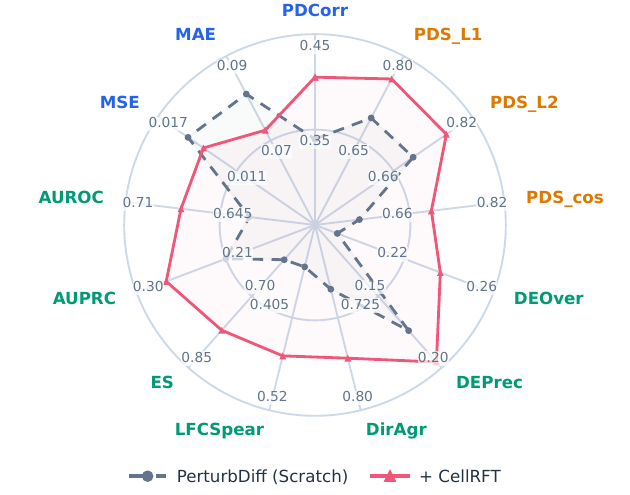}
  \end{subfigure}
  \captionsetup{font=footnotesize}
  \caption{\textbf{Training--evaluation objective mismatch.}
  \textbf{(a)} Standard pipelines optimize surrogate objectives, such as
  reconstruction and denoising, whose improvement need not yield higher
  biological fidelity. CellRFT instead uses biological evaluation feedback
  to guide model updates through reinforcement learning.
  \textbf{(b)} CellRFT fine-tunes
  PerturbDiff (Scratch)~\citep{yuan2026perturbdiff} on Replogle. The resulting
  model (+CellRFT) improves all 13 evaluated metrics, as defined in
  Table~\ref{tab:metric-overview}.}
  \label{fig:intro-overview}
\end{figure}

Learning directly from biological evaluation could reduce this mismatch,
but many metrics involve non-differentiable thresholds or rankings.
To use such feedback, reinforcement learning (RL) fine-tuning offers a
potential route without differentiating through the evaluator, with
successful applications in language
models~\citep{ouyang2022training,guo2025deepseekr1,srivastava2025technical} and diffusion-based image
generation~\citep{black2024training,liu2025flowgrpo,zheng2026diffusionnft}.
In perturbation prediction, metrics such as affected-gene recovery assess
responses across cells, so RL formulations requiring individual-output
scores cannot be transferred directly.
These metrics also assess distinct biological properties, so using them
jointly as rewards requires understanding whether optimizing one helps or
harms the others, a task-specific question that remains insufficiently
explored.

To address these challenges, we propose \textbf{CellRFT}, a reinforcement fine-tuning
framework that brings biological evaluation into single-cell perturbation
model optimization and examines how reward choices shape predictive
performance (Figure~\ref{fig:intro-overview-pipeline}). CellRFT uses policy-gradient RL to optimize biological rewards
without differentiating through the evaluation metrics.
To bridge the gap between individual generation trajectories and
population-level evaluation, it generates multiple candidate sets of cells
for each perturbation and shares each set's relative reward feedback
across its constituent trajectories.
To integrate multiple biological criteria, CellRFT uses hierarchical
reward aggregation, averaging first within evaluation categories and then
across categories to give each category equal total weight regardless of
its number of metrics.
We further investigate reward integration through systematic comparisons
of individual rewards, category combinations, and aggregation rules,
together with an analysis of how optimizing each reward affects other
evaluation criteria.
Comprehensive experiments demonstrate CellRFT's applicability across
different pretrained models and effectiveness in improving perturbation
prediction through direct biological feedback.
Figure~\ref{fig:intro-overview-radar} illustrates these gains for PerturbDiff
(Scratch), where CellRFT improves all 13 reported evaluation metrics.
The experiments further reveal that optimizing one biological criterion can help or
hinder others, while complementary rewards can jointly improve criteria
beyond those directly optimized.
Understanding these interactions can inform both the choice of training
rewards and the selection of criteria for evaluating biological predictions.

In summary, our contributions are threefold:
\begin{itemize}
  \item We introduce CellRFT, a reinforcement fine-tuning framework that
  turns biological evaluation into direct training feedback for single-cell
  perturbation models, addressing the mismatch between training objectives
  and biological evaluation criteria.
  \item We reveal how biological rewards reinforce or conflict with one
  another during optimization, showing that distinct evaluation criteria
  can benefit from shared training objectives and informing the design of
  complementary rewards.
  \item Comprehensive experiments demonstrate CellRFT's applicability across
  different pretrained models and effectiveness in improving perturbation
  prediction, and show how
  reward composition and aggregation can balance improvements across
  biological criteria.
\end{itemize}

\section{Related Work}
\label{sec:related-work}


\subsection{Single-Cell Perturbation Modeling}
\label{sec:related-perturbation}

Single-cell perturbation prediction seeks to infer post-perturbation
gene-expression profiles from control cells, a specified perturbation,
and the cellular context.
To represent the effect of a perturbation on cellular state,
scGen~\citep{lotfollahi2019scgen} learns shifts in a latent expression
space. CPA~\citep{lotfollahi2023predicting} further separates perturbation
and cellular covariates in that space, allowing their effects to be
recombined for prediction under new conditions.
For generalization to unseen genetic perturbations,
GEARS~\citep{roohani2024predicting} uses gene--gene relationships to
share information between observed and unobserved perturbation targets.
Beyond representing these conditional effects, learning from
single-cell measurements requires handling unpaired observations:
sequencing destroys the measured cells, so the same cell cannot be
observed before and after perturbation.
CellOT~\citep{bunne2023learning} addresses this by learning an optimal
transport map between control and perturbed populations, while
STATE~\citep{adduri2025predicting} learns responses using a
population-matching objective.
Diffusion and flow models provide flexible generative formulations
for these response distributions: Squidiff~\citep{he2025squidiff}
learns a denoising process, and CellFlow~\citep{klein2025cellflow}
learns conditional flows.
PerturbDiff~\citep{yuan2026perturbdiff} extends diffusion to the space
of cell distributions to capture variation among possible response
distributions under the same observed conditions.

Despite their different modeling strategies, these methods typically
optimize surrogate objectives, such as reconstruction, denoising, or
distribution matching, rather than the biological criteria used for
evaluation. This training--evaluation objective mismatch means that
improvements in the training loss need not translate into better
biological predictions, motivating direct use of biological evaluation
as training feedback.

\subsection{Reinforcement Learning and Biological Evaluation}
\label{sec:related-rl-evaluation}

Reinforcement learning (RL) has driven substantial progress across
diverse fields by incorporating evaluation feedback into training
and directing optimization toward task-level objectives.
In language-based assistance and reasoning, likelihood-based
training alone does not ensure that outputs follow user instructions,
solve problems correctly, or execute successfully as code.
RL introduces human preference judgments, answer verification, and
execution results as rewards, improving instruction following,
reasoning accuracy, and the completion of software engineering
tasks~\citep{ouyang2022training,guo2025deepseekr1,cao2026qwen3coder}.
In visual generation, denoising and flow-matching losses do not
directly measure whether images satisfy perceptual preferences
or accurately depict the objects and relations specified in a prompt.
Optimizing rewards derived from preference models and multimodal
evaluators improves visual quality and compositional consistency,
bringing training closer to the criteria used to judge generated
images~\citep{black2024training,liu2025flowgrpo,zheng2026diffusionnft}.
In molecular design, reproducing known chemical structures does
not ensure the activity and physicochemical properties needed
for a useful compound. Property-based RL rewards guide optimization
toward these requirements, enriching candidates for predicted
target activity and, when combined with synthesis constraints,
supporting the discovery of experimentally validated
antibiotics~\citep{olivecrona2017molecular,swanson2026synthemolrl}.

In perturbation modeling, biologically meaningful predictions
should capture overall transcriptional responses, recover the
genes affected by an intervention, and distinguish the effects
of different perturbations.
To assess these properties, prior studies have developed and adopted
complementary metrics of expression
fidelity~\citep{lotfollahi2019scgen,bunne2023learning},
differential-expression recovery~\citep{zhu2025auprc}, and
perturbation discrimination~\citep{wu2024perturbench,arc2025vcc}.
Although Cell-Eval, released alongside STATE, consolidates these
evaluation dimensions in a common suite~\citep{adduri2025predicting},
how the corresponding metrics should jointly guide model optimization
remains unresolved.
Using them as biological rewards requires understanding whether
optimizing one criterion reinforces or compromises performance
on the others.

The most closely related works, developed concurrently with ours, are
$D^{2}R^{2}$~\citep{fan2026d2r2} and
PerturbCellRL~\citep{wu2026perturbcellrl}.
Both seek to improve the biological consistency of predicted cells,
combining RL with regulatory relationships or pathway knowledge
to guide prediction. In contrast, CellRFT offers a complementary perspective by
studying two questions that remain insufficiently explored: how to
integrate commonly used Cell-Eval
metrics into RL training, and how their complementarity and redundancy
should guide the construction of a unified biological objective.

\section{Method}
\label{sec:method}

This section first introduces the problem formulation and biological
evaluation criteria, then presents CellRFT and its training procedure.

\subsection{Problem Formulation and Biological Evaluation}
\label{sec:problem-evaluation}

\paragraph{Problem Definition.}
For cellular context $c$ (e.g., cell type or batch) and perturbation
$\tau$, consider unpaired control population
$X_{\mathrm{ctrl}}\in\mathbb{R}^{N_{\mathrm{ctrl}}\times|\mathcal{G}|}$
and perturbed population
$X_{\mathrm{pert}}\in\mathbb{R}^{N_{\mathrm{pert}}\times|\mathcal{G}|}$,
with cells in rows and genes $\mathcal{G}$ in columns.
Perturbation modeling aims to learn a conditional generator
$p_\theta(\,\cdot\mid X_{\mathrm{ctrl}},c,\tau)$ whose predicted cell
distribution $\widehat P_{c,\tau}$ approximates the true perturbed-cell
distribution $P_{c,\tau}$, i.e., $\widehat P_{c,\tau}\approx P_{c,\tau}$,
while maximizing the expected biological reward of predictions
$\widehat X\in\mathbb{R}^{N\times|\mathcal{G}|}$:
\begin{equation}
  \max_{\theta}\;
  \mathbb{E}_{\widehat X\sim
    p_\theta(\,\cdot\mid X_{\mathrm{ctrl}},c,\tau)}
  \!\left[r\!\left(\widehat X;
    X_{\mathrm{ctrl}},X_{\mathrm{pert}},c,\tau\right)\right],
  \label{eq:reward-objective}
\end{equation}
where $r$ is a reward function given by the biological evaluation metric
itself when larger values indicate better performance, and by its negative
otherwise.

\paragraph{Biological Evaluation.}
Commonly used metrics are available in Cell-Eval, introduced alongside
STATE~\citep{adduri2025predicting}. These metrics assess agreement between
predicted and ground-truth cell populations from three complementary
perspectives: (i) \textit{global response fidelity}, which measures how well
predictions reproduce the overall expression response averaged across cells;
(ii) \textit{differential-expression fidelity}, which focuses on recovery of
significantly affected genes and their response patterns; and
(iii) \textit{perturbation discrimination}, which evaluates whether a predicted
response can be matched to the correct perturbation.
Table~\ref{tab:metric-overview} summarizes the 14 selected metrics; see
\citet{yuan2026perturbdiff} for further details.
Scores are computed per perturbation $\tau$ and averaged over the evaluated
perturbations $\mathcal{P}$ unless stated otherwise.

\begin{table}[!t]
  \centering
  \small
  \setlength{\tabcolsep}{4pt}
  \renewcommand{\arraystretch}{1.12}
  \caption{Biological evaluation metrics organized by evaluation perspective.
  Arrows indicate the preferred direction; ranges assume nondegenerate inputs.}
  \label{tab:metric-overview}
  \begin{NiceTabular}{@{}>{\centering\arraybackslash}p{0.16\linewidth}
                      >{\raggedright\arraybackslash}p{0.255\linewidth}
                      >{\raggedright\arraybackslash}p{0.105\linewidth}
                      >{\raggedright\arraybackslash}p{0.29\linewidth}c@{}}
    \toprule
    Perspective & Metric name & Abbrev. & Evaluation focus & Range \\
    \midrule
    \Block{3-1}{\textbf{Global}\\\textbf{response}\\\textbf{fidelity}}
      & Mean squared error & MSE & Squared error in average gene expression
      & $[0,\infty)\,\downarrow$ \\
      & Mean absolute error & MAE & Absolute error in average gene expression
      & $[0,\infty)\,\downarrow$ \\
      & Pearson delta correlation & PDCorr & Alignment of predicted and observed expression changes relative to controls
      & $[-1,1]\,\uparrow$ \\
    \midrule
    \Block{8-1}{\textbf{Differential-}\\\textbf{expression}\\\textbf{fidelity}}
      & Differential-expression overlap & DEOver & Overlap between predicted and observed DE genes
      & $[0,1]\,\uparrow$ \\
      & Differential-expression precision & DEPrec & Agreement of predicted DE genes with observed DE genes
      & $[0,1]\,\uparrow$ \\
      & Significant-DE-gene recall & DESigRec & Recovery of observed significant DE genes
      & $[0,1]\,\uparrow$ \\
      & Direction agreement & DirAgr & Agreement in up- and downregulation of observed DE genes
      & $[0,1]\,\uparrow$ \\
      & Log-fold-change Spearman correlation & LFCSpear & Rank agreement of predicted and observed log-fold changes in DE genes
      & $[-1,1]\,\uparrow$ \\
      & Area under the receiver operating characteristic curve & AUROC & Discrimination between DE and non-DE genes
      & $[0,1]\,\uparrow$ \\
      & Area under the precision--recall curve & AUPRC & Precision--recall performance in identifying observed DE genes
      & $[0,1]\,\uparrow$ \\
      & Effect size correlation & ES & Rank agreement of predicted and observed DE-gene counts across perturbations
      & $[-1,1]\,\uparrow$ \\
    \midrule
    \Block{3-1}{\textbf{Perturbation}\\\textbf{discrimination}}
      & Perturbation discrimination score & PDS$_{\mathrm{L1}}$ & Matching predictions to the correct perturbation using L1 distance
      & $[1/L,1]\,\uparrow$ \\
      & Perturbation discrimination score & PDS$_{\mathrm{L2}}$ & Matching predictions to the correct perturbation using L2 distance
      & $[1/L,1]\,\uparrow$ \\
      & Perturbation discrimination score & PDS$_{\mathrm{cos}}$ & Matching predictions to the correct perturbation using cosine distance
      & $[1/L,1]\,\uparrow$ \\
    \bottomrule
  \end{NiceTabular}
  \par\smallskip
  \begin{minipage}{\linewidth}
    \footnotesize
    DE: differential expression; $L=|\mathcal{P}|$: number of reference
    perturbations. Cell-Eval's significant-DE-gene recall is
    $\mathrm{DESigRec}_\tau
    =|\widehat{\mathcal{D}}_\tau\cap\mathcal{D}_\tau|/|\mathcal{D}_\tau|$,
    where $\mathcal{D}_\tau$ and $\widehat{\mathcal{D}}_\tau$ are the observed
    and predicted significant DE-gene sets, respectively, identified against
    controls~\citep{adduri2025predicting}.
  \end{minipage}
\end{table}

\subsection{CellRFT: Reinforcement Fine-Tuning with Biological Evaluation}
\label{sec:cellrft}

Perturbation modeling aims to maximize the expected biological reward
defined in Eq.~\eqref{eq:reward-objective}. In practice, models are typically
trained with differentiable surrogate losses, but reducing these losses
does not necessarily improve biological evaluation scores. Using biological
evaluation directly as training feedback presents three challenges:
(i) \textit{non-differentiability}: significance thresholds, rankings, and
gene-set operations in many metrics prevent direct backpropagation;
(ii) \textit{evaluation granularity}: biological metrics evaluate cell
populations, while existing sample-level reinforcement learning (RL)
formulations for generative models require a reward for each individual
sample; and
(iii) \textit{multi-reward integration}: biological evaluation metrics
remain largely unexplored as RL rewards, leaving unclear how each shapes
model behavior and how to integrate these rewards into a unified
training objective.

\begin{figure}[!t]
  \centering
  \includegraphics[width=\linewidth]{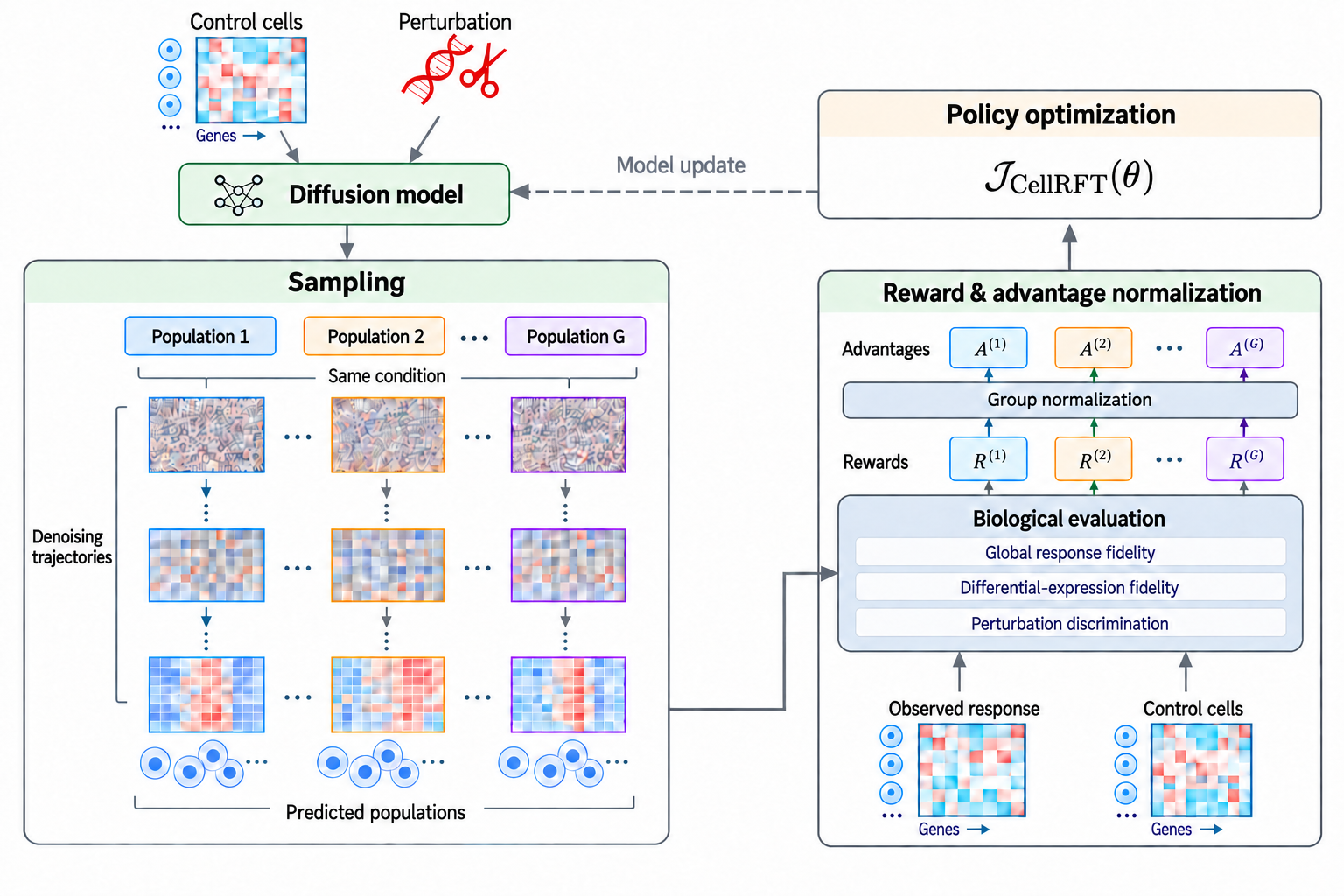}
  \captionsetup{font=footnotesize}
  \caption{\textbf{Overview of CellRFT.}
  A diffusion model samples $G$ candidate cell populations under the same
  condition. Biological evaluation compares each prediction with observed
  and control populations; metric rewards are averaged first within
  categories and then across categories to obtain $R^{(g)}$.
  Group normalization produces advantages $A^{(g)}$, each shared across
  all cells and denoising steps in its candidate population.
  These advantages guide policy optimization with
  $\mathcal J_{\mathrm{CellRFT}}(\theta)$
  (Eq.~\eqref{eq:cellrft-objective}) to update the model, completing the
  sampling--evaluation--optimization cycle.}
  \label{fig:method-framework}
\end{figure}

To handle these, we propose \textbf{CellRFT}, a reinforcement fine-tuning
framework for single-cell perturbation modeling. It handles
non-differentiability through policy-gradient RL, bridges the evaluation
granularity gap through population-level sampling and shared feedback,
and combines multiple rewards through hierarchical averaging.
Figure~\ref{fig:method-framework} summarizes the sampling,
reward and advantage computation, and policy optimization stages.
The details are as follows.

\paragraph{Reinforcement Learning from Evaluation Feedback.}
Indeed, RL fine-tuning has driven substantial progress in large language
models and text-to-image generation by incorporating evaluation into
training. We use Flow-GRPO~\citep{liu2025flowgrpo}, a representative
framework for image generation, to illustrate how RL bridges the
training--evaluation gap.

Flow-GRPO alternates between sampling candidate images, evaluating them,
and updating the generator using their relative performance.
In the sampling stage, the model generates a group of candidate images
$\{\widehat{\mathbf{x}}^{(g)}\}_{g=1}^{G}$ under the same text prompt.
Each image is produced by a \textit{trajectory} $\rho^{(g)}$, a sequence
of stochastic generation steps, and receives a scalar evaluation
\textit{reward} $r(\widehat{\mathbf{x}}^{(g)})$.
To account for differences in reward baselines and scales across prompts,
Flow-GRPO normalizes rewards within each prompt group:
\begin{equation}
  A^{(g)}=
  \frac{r(\widehat{\mathbf{x}}^{(g)})
    -\text{mean}(\{r(\widehat{\mathbf{x}}^{(h)})\}_{h=1}^{G})}
  {\text{std}(\{r(\widehat{\mathbf{x}}^{(h)})\}_{h=1}^{G})},
  \qquad g=1,\ldots,G,
  \label{eq:image-advantage}
\end{equation}
The resulting \textit{advantage} $A^{(g)}$ measures relative performance within the
group, reducing sensitivity of update weights to a prompt's absolute
reward level and dispersion.

These advantages guide the model update by favoring trajectories that
outperform others under the same prompt. The generator acts as a \textit{policy},
with $p_\theta(\rho)$ denoting a trajectory's probability density.
The reward-driven part of the local update follows the gradient of an
advantage-weighted log-likelihood surrogate:
\begin{equation}
  \nabla_\theta\!\left[
    \frac{1}{G}\sum_{g=1}^{G}
    A^{(g)}\log p_\theta(\rho^{(g)})
  \right]
  =\frac{1}{G}\sum_{g=1}^{G}
    A^{(g)}\nabla_\theta\log p_\theta(\rho^{(g)}).
  \label{eq:reward-policy-gradient}
\end{equation}
Here the prompt, sampled trajectories, and advantages are held fixed.
The gradient acts on the model's log likelihood, with $A^{(g)}$ supplying
the evaluation feedback without requiring derivatives of the evaluator.
Thus, non-differentiable evaluation can guide generation.

However, moving from image generation to perturbation modeling changes
the granularity of evaluation: Flow-GRPO evaluates individual images,
whereas biological evaluation assesses single-cell perturbation predictions
collectively over sets of cells.
Moreover, Flow-GRPO formulates optimization around a single
scalar reward, while integrating multiple biological evaluation criteria
into a unified reward remains an open research question.
Together, these challenges make Flow-GRPO's formulation not directly
applicable to perturbation modeling.

\paragraph{Population-Level Rewards and Hierarchical Aggregation.}
To incorporate population-level biological evaluation into training,
CellRFT treats a sampled set of generated cells as one candidate,
organizing sampling and reward comparison at the same granularity as
biological evaluation.

For each perturbation $\tau$, we first sample $G$ candidate sets of $B$ cells each:
\begin{samepage}
\begin{equation}
  \begin{aligned}
    \widehat X^{(g)}&=\{\widehat{\mathbf{x}}_n^{(g)}\}_{n=1}^{B},
    \qquad g=1,\ldots,G,\\
    \widehat{\mathbf{x}}_n^{(g)}
    &\sim p_\theta(\,\cdot\mid X_{\mathrm{ctrl}},c,\tau),
    \qquad n=1,\ldots,B.
  \end{aligned}
  \label{eq:population-rollouts}
\end{equation}
\end{samepage}
Then, we apply hierarchical aggregation to metric rewards, averaging
first within categories and then across categories:
\begin{equation}
  R^{(g)}
  =\frac{1}{|\mathcal C|}\sum_{k\in\mathcal C}
    \frac{1}{|\mathcal J_k|}\sum_{j\in\mathcal J_k}
    r_j\!\left(\widehat X^{(g)};
      X_{\mathrm{ctrl}},X_{\mathrm{pert}},c,\tau\right).
  \label{eq:population-reward}
\end{equation}
Here $\mathcal C$ indexes the three evaluation categories in
Table~\ref{tab:metric-overview}, and $\mathcal J_k$ contains the selected
reward metrics in category $k$. By hierarchically aggregating rewards, we give each category equal total weight, avoiding optimization bias toward categories with more metrics.

\paragraph{Policy Optimization with Shared Population Advantages.}
We next use the aggregated rewards to iteratively update the generator,
increasing the likelihood of candidate sets with higher biological rewards.

To account for differences in reward baselines and scales across
perturbations, we normalize rewards within each perturbation
group~\citep{liu2025flowgrpo} to obtain a relative advantage for each candidate:
\begin{equation}
  A^{(g)}=
  \frac{R^{(g)}-\text{mean}(\{R^{(h)}\}_{h=1}^{G})}
  {\text{std}(\{R^{(h)}\}_{h=1}^{G})}.
  \label{eq:population-advantage}
\end{equation}
Each $A^{(g)}$ is shared across all denoising steps of the candidate's
$B$ cells.

Let $\mathbf{x}_{n,t}^{(g)}$ denote the state of cell $n$ in candidate $g$
at step $t$ of a $T$-step denoising trajectory, with
$\mathbf{x}_{n,0}^{(g)}$ the generated cell. Using the shared advantages,
CellRFT maximizes the following clipped objective~\citep{liu2025flowgrpo}:
\begin{equation}
  \mathcal J_{\mathrm{CellRFT}}(\theta)
  =\mathbb E\Bigg[\frac{1}{GBT}
    \sum_{g=1}^{G}\sum_{n=1}^{B}\sum_{t=1}^{T}
    \Big\{\min\!\Big(w_{n,t}^{(g)}A^{(g)},
    \operatorname{clip}\!\left(w_{n,t}^{(g)},1-\eta,1+\eta\right)
    A^{(g)}\Big)
    \Big\}\Bigg],
  \label{eq:cellrft-objective}
\end{equation}
where the expectation is over training conditions and candidate
trajectories sampled from the old policy $p_{\theta_{\mathrm{old}}}$.
The ratio $w_{n,t}^{(g)}$ compares the current denoising transition density
with that of the sampling policy:
\begin{equation}
  w_{n,t}^{(g)}(\theta)
  =\frac{p_\theta(\mathbf{x}_{n,t-1}^{(g)}\mid
    \mathbf{x}_{n,t}^{(g)},X_{\mathrm{ctrl}},c,\tau)}
  {p_{\theta_{\mathrm{old}}}(\mathbf{x}_{n,t-1}^{(g)}\mid
    \mathbf{x}_{n,t}^{(g)},X_{\mathrm{ctrl}},c,\tau)}.
  \label{eq:transition-ratio}
\end{equation}
Clipping with threshold $\eta$ limits the incentive to change this ratio
in the direction favored by $A^{(g)}$.

The updated generator then samples new candidate sets for evaluation,
and this sampling--evaluation--update cycle repeats throughout fine-tuning.
Algorithm~\ref{alg:cellrft} summarizes the complete training procedure.

\begin{algorithm}[!t]
\caption{CellRFT Training}
\label{alg:cellrft}
\begin{algorithmic}[1]
\Require Pretrained model $p_\theta$; training set of
         $(X_{\mathrm{ctrl}},X_{\mathrm{pert}},c,\tau)$ tuples;
         reward functions $r_j$
\Ensure Fine-tuned model $p_\theta$
\For{iteration $e=1,\ldots,E$}
    \State $\theta_{\mathrm{old}}\gets\theta$
    \State Sample a minibatch of $(X_{\mathrm{ctrl}},X_{\mathrm{pert}},c,\tau)$ tuples
    \ForAll{$(X_{\mathrm{ctrl}},X_{\mathrm{pert}},c,\tau)$ in the minibatch}
        \State Sample $G$ sets of $B$ cells and their trajectories from
               $p_{\theta_{\mathrm{old}}}$ (Eq.~\eqref{eq:population-rollouts})
        \State Compute hierarchically aggregated rewards $R^{(g)}$
               (Eq.~\eqref{eq:population-reward})
        \State Compute group-relative advantages $A^{(g)}$
               (Eq.~\eqref{eq:population-advantage})
    \EndFor
    \State Update $\theta$ using shared population advantages
           by maximizing Eq.~\eqref{eq:cellrft-objective}
\EndFor
\State \Return $p_\theta$
\end{algorithmic}
\end{algorithm}

\section{Experiments}
\label{sec:experiments}

This section evaluates CellRFT's applicability across different pretrained
models and effectiveness in improving perturbation prediction, then examines
how reward composition, aggregation, hierarchy, and complementarity shape
improvements across biological criteria.


\subsection{Experimental Setup}
\label{sec:experimental-setup}

\paragraph{Dataset and Evaluation Protocol.}
We evaluate on Replogle~\citep{nadig2025transcriptome}, following the
experimental protocol of \citet{yuan2026perturbdiff} with identical
preprocessing and data splits.
The benchmark covers 2,023 genetic perturbations across four cell lines
(K562, RPE1, Jurkat, and HepG2), with predictions evaluated on 2,000
highly variable genes. HepG2 is the target cell line: training uses
the other three cell lines together with 30\% of HepG2 perturbations,
while evaluation follows the original validation and test partitions.
Preprocessing includes knockdown-efficacy
filtering inherited from STATE~\citep{adduri2025predicting}, library-size
normalization, log transformation, and division of expression values by 10.
Further details are provided in \citet{yuan2026perturbdiff}.

\paragraph{Baselines and Evaluation Metrics.}
We compare against simple statistical baselines,
Mean~\citep{kernfeld2025comparison} and Linear~\citep{ahlmann2025deep}.
Mean averages expression within each perturbation; its three variants,
Mean Variant (per Cell Type), Mean Variant (per Batch), and
Mean Variant (Overall), instead average within each cell type, within
each experimental batch, and across all cells, respectively.
We also include representative perturbation models:
CPA~\citep{lotfollahi2023predicting}, STATE~\citep{adduri2025predicting},
CellFlow~\citep{klein2025cellflow}, Squidiff~\citep{he2025squidiff}, PerturbDiff (Scratch)~\citep{yuan2026perturbdiff}
and PerturbDiff (Finetuned)~\citep{yuan2026perturbdiff}.
We evaluate performance using 14 representative metrics from
Cell-Eval~\citep{adduri2025predicting},
covering global response fidelity, differential-expression fidelity,
and perturbation discrimination (Table~\ref{tab:metric-overview}).

\paragraph{Implementation Details.}
We apply CellRFT to both PerturbDiff (Scratch) and PerturbDiff
(Finetuned)~\citep{yuan2026perturbdiff}, corresponding to supervised training
from scratch and supervised fine-tuning after pretraining, respectively.
To maintain comparable reward scales and per-perturbation feedback,
we select 11 metrics from Table~\ref{tab:metric-overview}, excluding
the unbounded MSE and MAE and the cross-perturbation metric ES.
Using these rewards, we fine-tune with a learning rate of $10^{-5}$
and $G=6$ candidate sets per perturbation, each containing $B=32$ cells.

\subsection{Applicability and Effectiveness of CellRFT Across Pretrained Models}
\label{sec:experimental-results}

Table~\ref{tab:replogle-main} compares statistical and learned baselines
with the two PerturbDiff models before and after CellRFT on 13 Replogle
metrics.
\begin{table}[!t]
  \centering
  \footnotesize
  \setlength{\tabcolsep}{2.5pt}
  \renewcommand{\arraystretch}{1.15}
  \caption{Perturbation prediction performance on Replogle.
  Arrows indicate better performance, and bold highlights the best result
  for each metric within each block.}
  \label{tab:replogle-main}
  \begin{tabular*}{\linewidth}{@{\extracolsep{\fill}}l*{7}{r}@{}}
    \toprule
    \multicolumn{8}{c}{\textbf{Differential-expression fidelity}} \\
    \midrule
    Model & DEOver $\uparrow$ & DEPrec $\uparrow$ & DirAgr $\uparrow$ & LFCSpear $\uparrow$ & AUROC $\uparrow$ & AUPRC $\uparrow$ & ES $\uparrow$ \\
    \midrule
    Mean & 0.127 & 0.094 & 0.532 & 0.077 & 0.467 & 0.092 & 0.417 \\
    Mean Variant (per Cell Type) & 0.178 & 0.087 & 0.743 & 0.492 & 0.404 & 0.078 & 0.417 \\
    Mean Variant (per Batch) & 0.110 & 0.092 & 0.475 & -0.024 & 0.450 & 0.088 & 0.421 \\
    Mean Variant (Overall) & 0.111 & 0.093 & 0.474 & -0.027 & 0.453 & 0.089 & 0.416 \\
    Linear & 0.068 & 0.074 & 0.535 & 0.056 & 0.435 & 0.085 & 0.353 \\
    CPA & 0.173 & 0.087 & 0.746 & 0.499 & 0.401 & 0.081 & 0.637 \\
    STATE & \textbf{0.196} & \textbf{0.193} & \textbf{0.778} & \textbf{0.506} & \textbf{0.633} & \textbf{0.239} & \textbf{0.818} \\
    CellFlow & 0.110 & 0.093 & 0.472 & -0.033 & 0.434 & 0.086 & 0.442 \\
    Squidiff & 0.039 & 0.091 & 0.432 & 0.000 & 0.366 & 0.079 & 0.419 \\
    \midrule
    PerturbDiff (Scratch) & 0.190 & 0.174 & 0.702 & 0.342 & 0.625 & 0.210 & 0.623 \\
    CellRFT + PerturbDiff (Scratch) & \textbf{0.236} & \textbf{0.196} & \textbf{0.758} & \textbf{0.452} & \textbf{0.672} & \textbf{0.270} & \textbf{0.771} \\
    \midrule
    PerturbDiff (Finetuned) & 0.214 & 0.192 & 0.723 & 0.388 & 0.652 & 0.257 & 0.762 \\
    CellRFT + PerturbDiff (Finetuned) & \textbf{0.239} & \textbf{0.201} & \textbf{0.752} & \textbf{0.455} & \textbf{0.681} & \textbf{0.273} & \textbf{0.771} \\
    \bottomrule
  \end{tabular*}

  \par\medskip
  \begin{tabular*}{\linewidth}{@{\extracolsep{\fill}}l*{6}{r}@{}}
    \toprule
    \multicolumn{7}{c}{\textbf{Global response fidelity and perturbation discrimination}} \\
    \midrule
    Model & MSE $\downarrow$ & MAE $\downarrow$ & PDCorr $\uparrow$ & PDS$_{\mathrm{L1}}$ $\uparrow$ & PDS$_{\mathrm{L2}}$ $\uparrow$ & PDS$_{\mathrm{cos}}$ $\uparrow$ \\
    \midrule
    Mean & 0.0990 & 0.206 & 0.048 & 0.582 & 0.599 & 0.608 \\
    Mean Variant (per Cell Type) & 0.0085 & 0.057 & 0.412 & 0.502 & 0.505 & 0.502 \\
    Mean Variant (per Batch) & 0.0699 & 0.174 & 0.002 & 0.505 & 0.505 & 0.506 \\
    Mean Variant (Overall) & 0.0711 & 0.175 & -0.001 & 0.502 & 0.503 & 0.501 \\
    Linear & 0.0130 & 0.074 & 0.058 & 0.519 & 0.517 & 0.667 \\
    CPA & 0.0750 & \textbf{0.054} & 0.418 & 0.582 & 0.581 & 0.521 \\
    STATE & \textbf{0.0064} & 0.055 & \textbf{0.437} & \textbf{0.788} & \textbf{0.802} & \textbf{0.676} \\
    CellFlow & 0.0949 & 0.205 & -0.003 & 0.507 & 0.507 & 0.495 \\
    Squidiff & 4.6410 & 2.077 & 0.089 & 0.500 & 0.501 & 0.501 \\
    \midrule
    PerturbDiff (Scratch) & 0.0147 & 0.081 & 0.340 & 0.690 & 0.700 & 0.575 \\
    CellRFT + PerturbDiff (Scratch) & \textbf{0.0135} & \textbf{0.072} & \textbf{0.405} & \textbf{0.759} & \textbf{0.768} & \textbf{0.697} \\
    \midrule
    PerturbDiff (Finetuned) & \textbf{0.0146} & 0.079 & 0.376 & \textbf{0.758} & \textbf{0.762} & \textbf{0.639} \\
    CellRFT + PerturbDiff (Finetuned) & 0.0149 & \textbf{0.076} & \textbf{0.409} & 0.742 & 0.748 & 0.611 \\
    \bottomrule
  \end{tabular*}
\end{table}

The improvements in both PerturbDiff variants demonstrate CellRFT's
applicability across different pretrained models and effectiveness in
improving perturbation prediction. We refer to
CellRFT applied to PerturbDiff (Scratch) as the scratch variant, and
CellRFT applied to PerturbDiff (Finetuned) as the finetuned variant.
The scratch variant improves all 13 reported metrics over PerturbDiff
(Scratch). It also exceeds PerturbDiff (Finetuned) on every metric,
although PerturbDiff (Finetuned) first uses additional perturbation data
for pretraining before supervised fine-tuning. This result suggests that
RL feedback can further extract useful capacity from a model trained
without those extra pretraining data, making CellRFT useful for new cell
types or perturbation panels where large-scale pretraining data are
expensive to collect. The finetuned variant further improves nine
metrics over PerturbDiff (Finetuned), showing that CellRFT also adds
value after supervised fine-tuning.

The strongest gains concentrate on metrics that evaluate
perturbation-induced gene-expression responses. In the scratch variant,
DEOver increases from 0.190 to 0.236, LFCSpear from 0.342 to 0.452,
AUROC from 0.625 to 0.672, and AUPRC from 0.210 to 0.270. The finetuned
variant shows the same pattern, with consistent improvements across the
differential-expression metrics. Consequently, both variants outperform
all non-CellRFT baselines on DEOver, DEPrec, AUROC, and AUPRC. These
metrics directly test whether the model identifies the affected genes
and ranks their response magnitudes correctly, showing that CellRFT
improves the perturbation-specific signal central to biological
prediction rather than only matching average expression profiles.

The results also show that reward effects are coupled rather than
metric-local. Some improvements transfer to metrics that are not used
as rewards: MSE, MAE, and ES are excluded from training, yet all three
improve for the scratch variant, and MAE and ES also improve for the
finetuned variant. At the same time, including a metric in the joint
reward does not guarantee improvement: the finetuned variant gains on
differential-expression metrics but loses on all three PDS metrics.
These heterogeneous responses motivate an ablation study of reward design:
how individual metrics affect performance across the evaluation suite, and
how combining rewards changes the resulting trade-offs.

\begin{figure}[t]
  \centering
  \includegraphics[width=\linewidth]{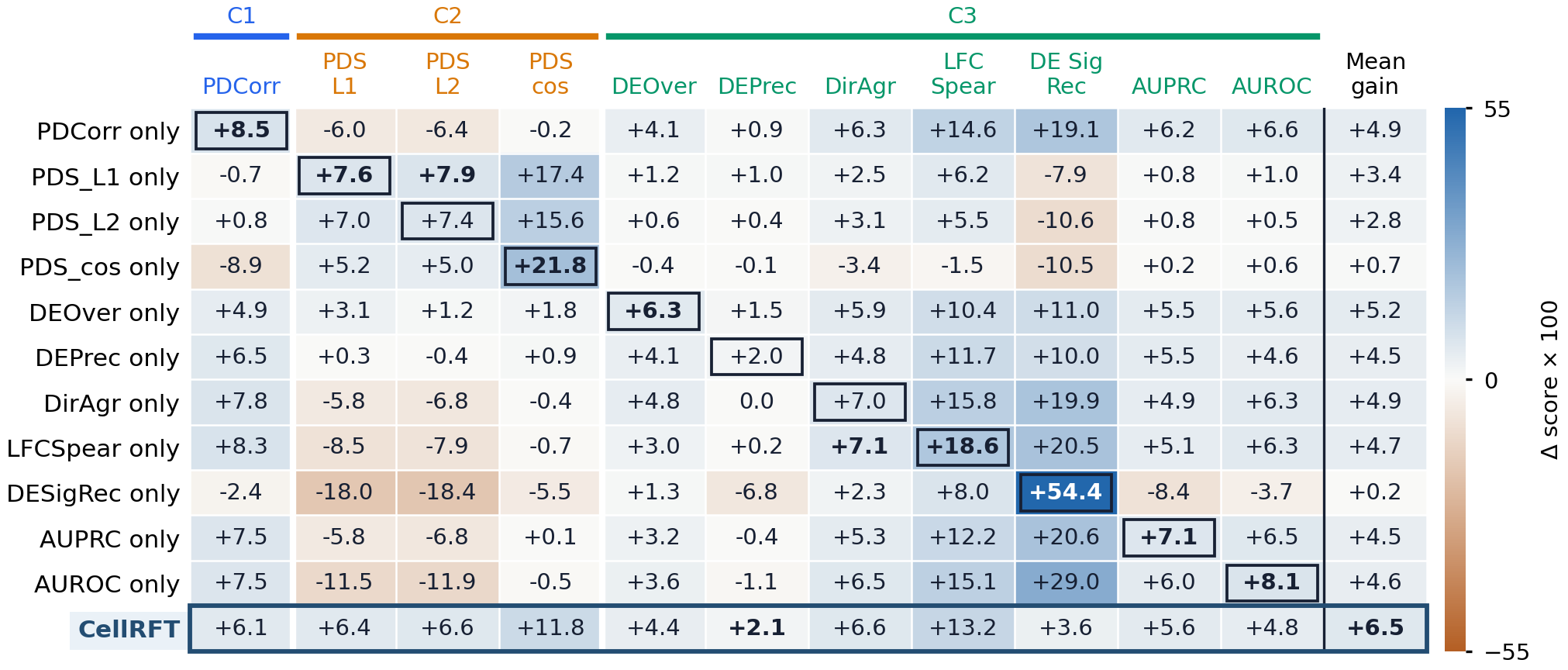}
  \caption{Cross-metric effects of single-reward fine-tuning. The ablation
  evaluates how using individual metrics as rewards affects performance
  across the evaluation suite, reporting gains over the pretrained model
  ($\times 100$). Single-reward variants use the corresponding evaluation
  metric as the sole reward, while CellRFT uses the hierarchical reward
  aggregation defined in Eq.~\eqref{eq:population-reward}. All single-reward
  variants improve their target metrics, but most incur cross-metric
  degradation; CellRFT achieves a more
  favorable overall trade-off, improving all 11 metrics with the highest
  mean gain.}
  \label{fig:single-reward}
\end{figure}

\subsection{Reward Composition and Aggregation}
The main results show that gains in one biological criterion can coexist
with losses in another. To test whether RL fine-tuning with a single
evaluation metric as the reward reproduces these trade-offs and whether
combining multiple metrics as rewards mitigates them, we design two
ablations: \textit{single-reward fine-tuning} to assess target and cross-metric
effects, and \textit{category-level combinations} to assess reward composition,
including a comparison of aggregation rules using all 11 rewards. Both ablations use the 11 reward metrics from
Table~\ref{tab:metric-overview}, grouped into global response fidelity
(C1: PDCorr), perturbation discrimination (C2: three PDS variants), and
differential-expression fidelity (C3: seven metrics). Single-reward
variants share the pretrained initialization, data split, and training
budget. Category-level variants use all reward metrics from one category
or a pair of categories; the full-reward comparison uses either an
unweighted arithmetic mean (Flat mean) or hierarchical averaging
(CellRFT; Eq.~\eqref{eq:population-reward}). Evaluation covers all 11
metrics, reporting gains over the pretrained model ($\times 100$) and
their arithmetic mean in the original metric scales.

\begin{figure}[t]
  \centering
  \includegraphics[width=\linewidth]{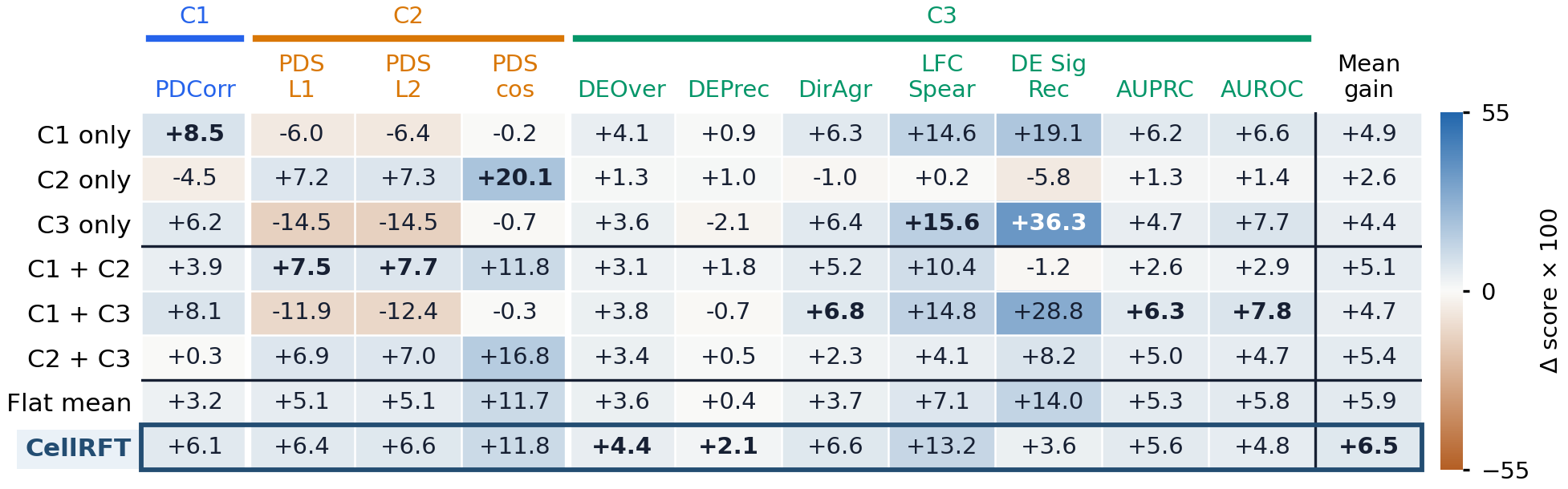}
  \caption{Cross-metric effects of reward composition and aggregation.
  The ablation evaluates how reward composition and aggregation affect
  performance across the evaluation suite, reporting gains over the
  pretrained model ($\times 100$). Category-level variants use all reward
  metrics from one evaluation category or a pair of categories. Flat mean
  takes the unweighted arithmetic mean of all 11 rewards, whereas CellRFT
  averages these rewards first within categories and then across categories
  (Eq.~\eqref{eq:population-reward}). Combining complementary reward
  categories mitigates cross-metric trade-offs, and CellRFT achieves the
  highest mean gain while improving all 11 metrics.}
  \label{fig:reward-composition}
\end{figure}

As shown in Figures~\ref{fig:single-reward}
and~\ref{fig:reward-composition}, we observe that:
(1) RL fine-tuning with each individual metric as the reward improves its
target metric, but most single-reward variants degrade other criteria,
with DESigRec showing the clearest divergence between target improvement
and cross-metric performance and DEOver improving all 11 metrics;
(2) combining complementary reward categories can mitigate these
trade-offs: fine-tuning with C3 alone reduces perturbation discrimination,
while fine-tuning with C2 alone reduces DESigRec, whereas C2+C3 is the only
pairwise category combination that improves all 11 metrics; and
(3) CellRFT improves all 11 metrics with the highest mean gain ($6.5$),
exceeding Flat mean ($5.9$) and DEOver-only ($5.2$), indicating a more
favorable overall trade-off despite not achieving the largest gain on
every metric.

\subsection{Reward Hierarchy and Complementarity}
\label{sec:reward-selection}

Although the single-reward and category-combination ablations reveal
cross-metric gains and trade-offs, they leave unresolved how these effects
are organized within and across the predefined evaluation categories in
Table~\ref{tab:metric-overview}. This structure matters for understanding
how CellRFT jointly improves multiple aspects of biological fidelity:
rewards from different categories may produce overlapping effects, whereas
rewards within the same category may contribute differently to shared
improvements.

\begin{figure}[!t]
  \centering
  \captionsetup[subfigure]{font=small,labelfont=bf,
    justification=raggedright,singlelinecheck=false,skip=3pt}
  \begin{subfigure}[b]{0.48\linewidth}
    \centering
    \includegraphics[width=\linewidth,height=8.255cm,keepaspectratio]{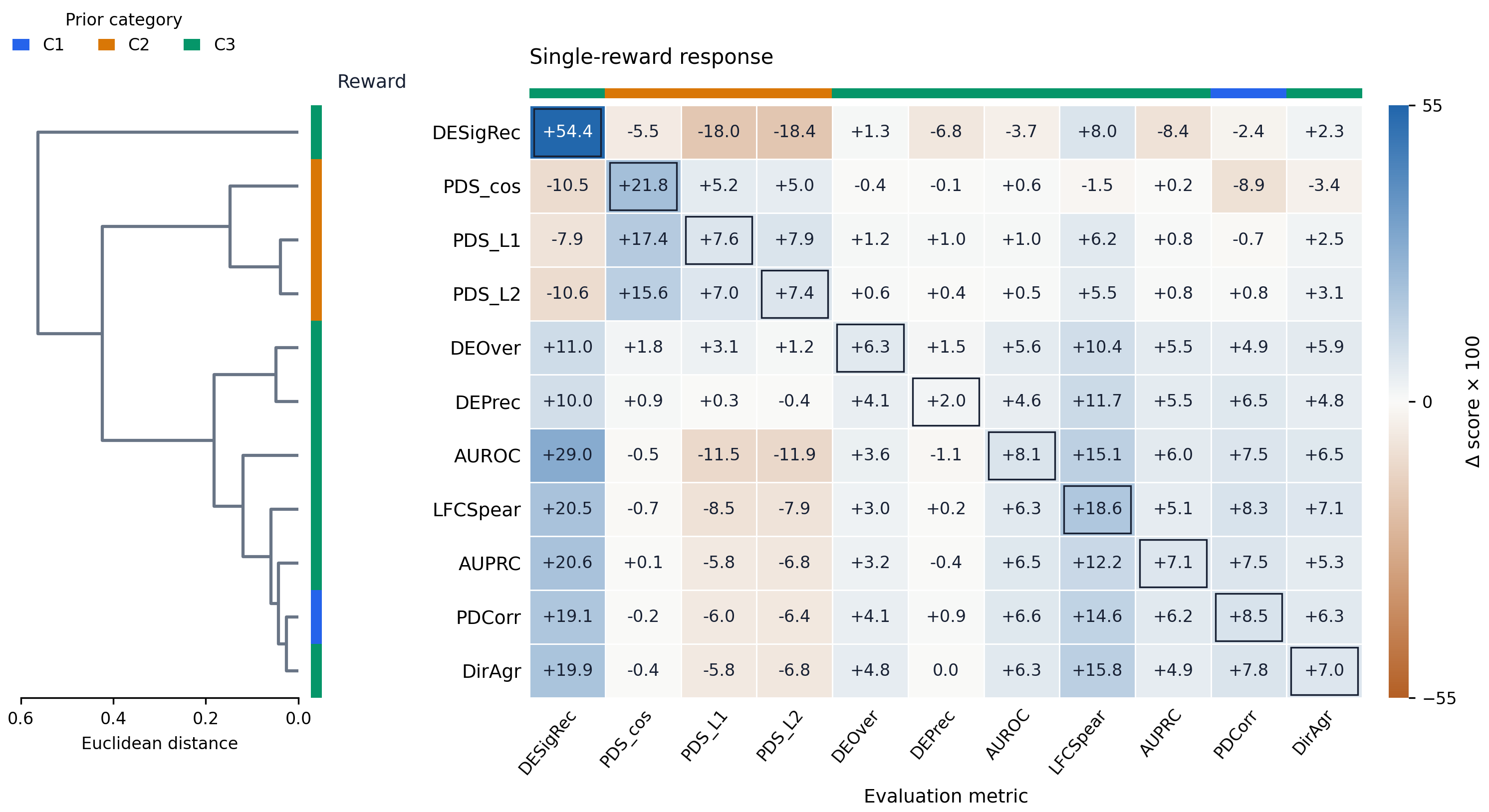}
    \caption{Hierarchical structure of reward responses.}
    \label{fig:reward-selection-hierarchy}
  \end{subfigure}\hfill
  \begin{subfigure}[b]{0.5\linewidth}
    \centering
    \includegraphics[width=8cm,height=4.5cm,keepaspectratio]{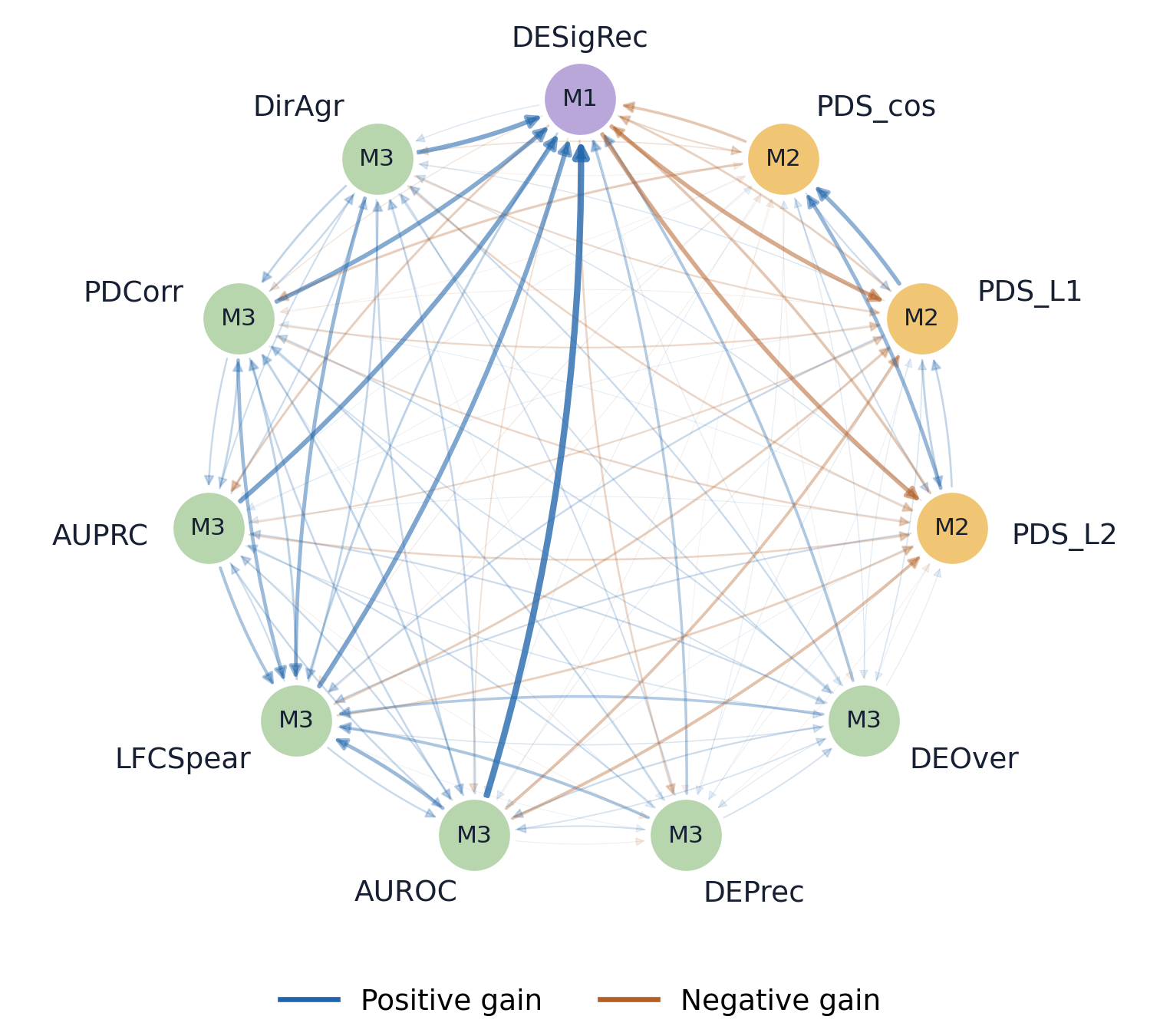}
    \caption{Community structure of metric interactions.}
    \label{fig:reward-selection-network}
  \end{subfigure}

  \medskip
  \begin{subfigure}[b]{0.48\linewidth}
    \centering
    \includegraphics[width=\linewidth,height=8.338cm,keepaspectratio]{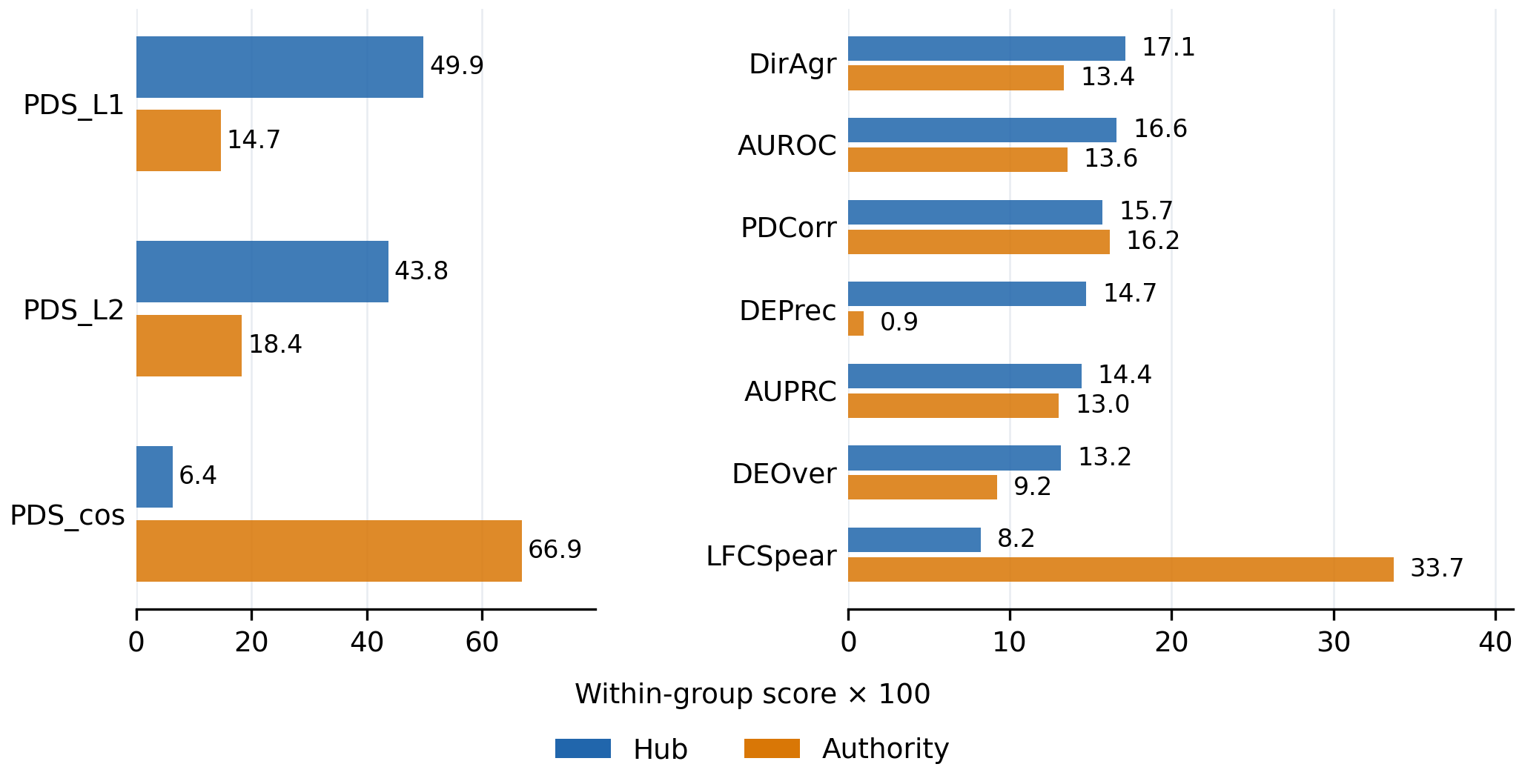}
    \caption{Influence and support within communities.}
    \label{fig:reward-selection-hits}
  \end{subfigure}\hfill
  \begin{subfigure}[b]{0.5\linewidth}
    \centering
    \includegraphics[width=8cm,height=4.5cm,keepaspectratio]{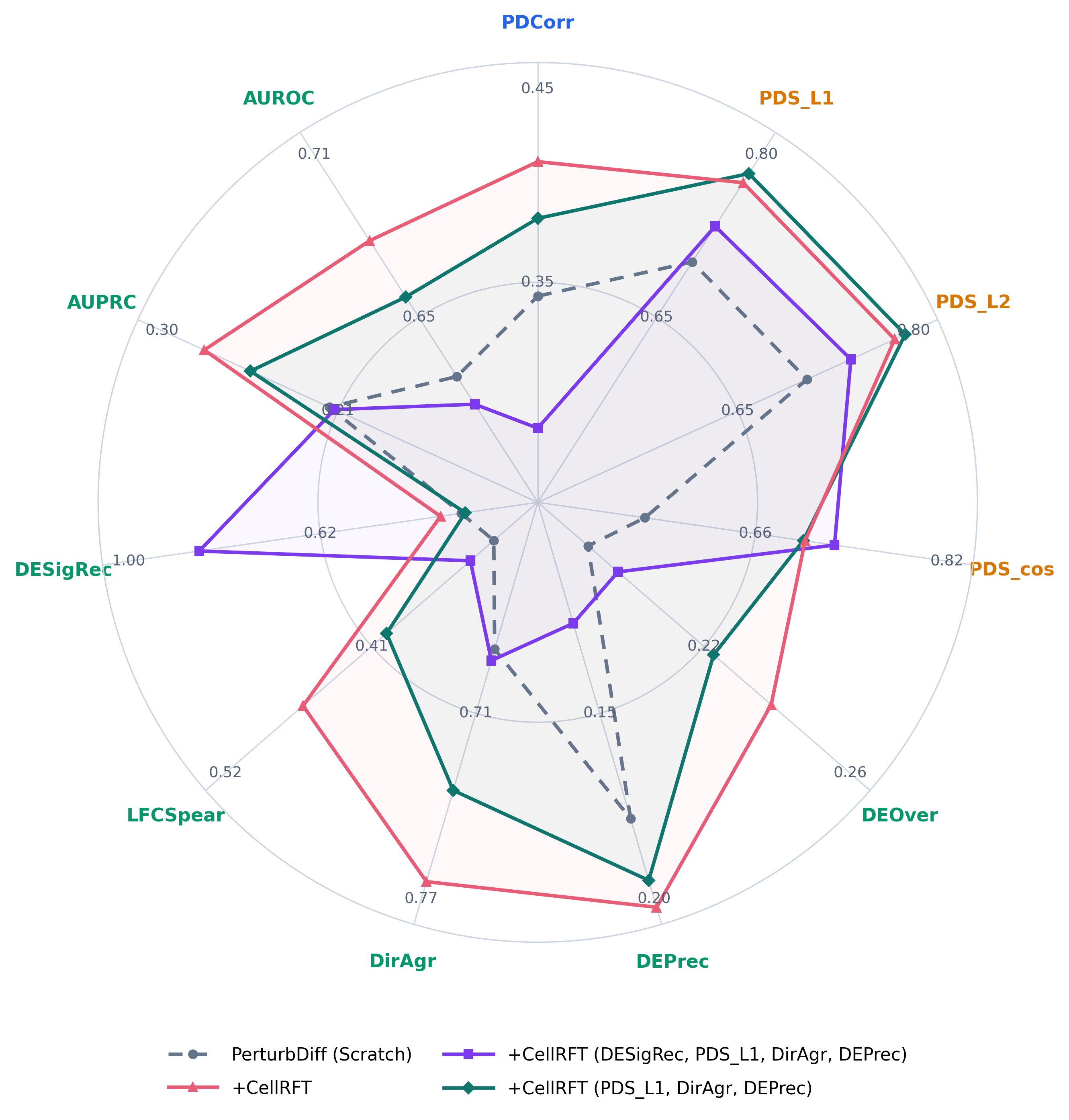}
    \caption{Fine-tuning with reward subsets.}
    \label{fig:reward-selection-radar}
  \end{subfigure}

  \captionsetup{font=footnotesize}
  \caption{\textbf{Hierarchical structure and complementarity of rewards.}
  \textbf{(a)} Average-linkage clustering of single-reward response
  profiles using Euclidean distances reveals similarities across prior
  evaluation categories, with PDCorr and DirAgr forming the closest pair.
  \textbf{(b)} A directed network of cross-metric gains, colored by the
  three communities identified from the hierarchy, reveals asymmetric
  support and trade-offs within and across communities.
  \textbf{(c)} Within-community HITS analysis of positive gains distinguishes
  influential rewards such as PDS\_L1 from supported metrics such as
  PDS\_cos, while highlighting limited indirect support for DEPrec.
  \textbf{(d)} Joint fine-tuning with three core rewards improves 10 of 11
  metrics over Scratch, showing that a compact subset can yield broad
  gains, while CellRFT with all 11 rewards provides a superior overall trade-off.}
  \label{fig:reward-selection}
\end{figure}

To this end, we first represent each single-reward variant by its gains over the
pretrained model across all 11 metrics and cluster these raw response
vectors using average linkage and Euclidean distances
(Figure~\ref{fig:reward-selection}a).
The resulting hierarchy partly recovers the predefined evaluation
categories while revealing cross-category similarities in optimization
effects. Specifically, the three PDS variants form a common branch,
consistent with their shared perturbation-discrimination task, whereas DESigRec separates from
the other differential-expression metrics. More notably, PDCorr and DirAgr
form the closest pair despite belonging to different evaluation categories:
both produce similar declines in perturbation-discrimination metrics and
similar gains across the remaining metrics.

We then examine how rewards support one another through a directed network,
where an edge from $i$ to $j$ records the gain in metric $j$ after optimizing
reward $i$ (Figure~\ref{fig:reward-selection}b). For network visualization
and within-group analysis, we use a three-group cut, supported by the
highest mean silhouette score ($0.624$) among $k=2,\ldots,10$ and the
largest jump in successive merge distances. We apply Hyperlink-Induced
Topic Search (HITS)~\citep{kleinberg1999authoritative}, a network-ranking
method, to positive, non-self gains within each non-singleton group.
Hub scores characterize rewards providing support, while authority scores
characterize metrics receiving it. This analysis reveals that similar
response profiles can conceal distinct optimization roles. Specifically,
PDS\_L1 improves PDS\_cos more strongly than the reverse, with a high hub
score for PDS\_L1 and a high authority score for PDS\_cos
(Figure~\ref{fig:reward-selection}c). Meanwhile, DEPrec receives limited
within-group support despite clustering closely with DEOver: similarity
in the effects a reward produces does not imply strong support for its
own metric from other rewards. DirAgr also emerges as a leading hub
candidate in the larger group. These roles suggest considering both
rewards that improve other metrics and metrics that receive limited
improvement from optimizing other rewards.

We select the metrics with the highest hub scores in the two non-singleton
groups (PDS\_L1 and DirAgr), retain DESigRec from the singleton group, and
add DEPrec to supplement limited precision support. We jointly fine-tune
using hierarchical averaging (Eq.~\eqref{eq:population-reward}) over the
selected response groups. Joint training reveals both broad indirect
improvements and a recall-driven imbalance (Figure~\ref{fig:reward-selection}d).
The four-metric subset exhibits a pattern consistent with reward hacking:
recall rises from $0.382$ to $0.833$, accompanied by precision and PDCorr
falling below the pretrained baseline. Removing DESigRec and reaveraging
the retained groups yields a recall-excluded variant that improves 10 of
11 evaluation metrics using only three direct rewards.
The representative subset demonstrates that a few direct objectives can
drive improvements across multiple biological criteria. Full-reward
CellRFT further achieves a more favorable balance across these criteria.
These findings show that distinct evaluation criteria need not each
require a separate training objective, providing an empirical basis for
selecting rewards through their ability to support broader improvements.

Overall, these experiments demonstrate CellRFT's applicability across
different pretrained models and effectiveness in improving perturbation
prediction, with particularly strong gains in differential-expression fidelity.
Reward ablations show that improvements in individual metrics can incur
losses elsewhere, and that reward composition and aggregation are important
for balancing these effects. Response analysis and joint fine-tuning
further demonstrate that a few representative objectives can support broad
improvements across biological criteria, while the full reward set achieves
a more favorable overall trade-off.

\section{Conclusion}
\label{sec:conclusion}

We introduced CellRFT to address the mismatch between surrogate training
objectives and biological evaluation by using evaluation feedback for
reinforcement fine-tuning.
Our analysis shows that biological rewards can reinforce or conflict with
one another, and that complementary objectives can improve criteria beyond
those directly optimized.
Comprehensive experiments demonstrate CellRFT's applicability across
different pretrained models and effectiveness in improving perturbation
prediction.

The trade-offs exposed by reward optimization show why a gain in one
metric need not mean a better biological prediction.
Consensus on which metrics reliably establish such improvement remains
incomplete, as illustrated by the revised evaluation criteria of the 2025 ~\citep{arc2025vcc}
and 2026 ~\citep{adduri2026virtual} Virtual Cell Challenges.
We hope CellRFT will inform metric selection by revealing what optimizing
each criterion improves or compromises, helping the community rethink
how progress in virtual cell modeling should be measured.

\clearpage
\bibliographystyle{plainnat}
\bibliography{references}

\end{document}